\documentclass[conference]{IEEEtran}
\IEEEoverridecommandlockouts
\usepackage{cite}
\usepackage{amsmath,amssymb,amsfonts}
\usepackage{algorithmic}
\usepackage{graphicx}
\usepackage{textcomp}
\usepackage{xcolor}
\def\BibTeX{{\rm B\kern-.05em{\sc i\kern-.025em b}\kern-.08em
    T\kern-.1667em\lower.7ex\hbox{E}\kern-.125emX}}
\begin{document}

\title{Self-Adaptive Systems in Organic Computing: Different Concepts of Self-Improvement\\
}

\author{\IEEEauthorblockN{Andreas Niederquell
\\
Seminar "Intelligent Systems"
\\
(Prof. Dr.-Ing. Sven Tomforde)
\\
University of Passau}
}

\maketitle

\begin{abstract}
With the intensified use of intelligent things, the demands on the technological systems are increasing permanently. A possible approach to meet the continuously changing challenges is to shift the system integration from design to run-time by using adaptive systems. Diverse adaptivity properties -- so-called self-* properties -- form the basis of these systems and one of the properties is self-improvement. It describes the ability of a system not only to adapt to a changing environment according to a predefined model, but also the capability to adapt the adaptation logic of the whole system.
In this paper, a closer look is taken at the structure of self-adaptive systems. Additionally, the systems' ability to improve themselves during run-time is described from the perspective of Organic Computing. Furthermore, four different strategies for self-improvement are presented, following the taxonomy of self-adaptation suggested by Christian Krupitzer et al.

\end{abstract}


\section{Introduction}
In 1991, the computer scientist Mark Weiser formulated his vision of \textit{Ubiquitous Computing}, predicting that in the 21\textsuperscript{st} century the personal computers would be replaced by \textit{intelligent things} \cite{b1}. The times are long gone, where there was at most one computer for each person. Nowadays we are surrounded by computer systems and devices which are contained in our everyday objects like mobile phones, cars, televisions or even refrigerators. But not only has the mere amount of the used intelligent things been increasing for the last 20 years; with the intensified use of technology the demands on these systems have also been increasing and will still continue to rise. The main problem occurs when it comes to the interconnection between the many different systems.

A suggestion for facing this problem was contributed in 2003 in another vision by the IBM researchers Jeffrey O. Kephart and David M. Chess \cite{b2}. In their research paper "The Vision of Autonomic Computing" they introduced the term \textit{Autonomic Computing} as the ability of computer systems "to manage themselves given high-level objectives from administrators". Inspired by the self-governance of social and economic systems, as well as the autonomic nervous system, the researchers invented the concept of Autonomic Computing as a long term challenge containing several important milestones along the path to a usable self-management of computer systems. According to Kephart and Chess, the journey toward fully autonomic computing would begin with automated functions that collect and aggregate information to support decisions made by human administrators. With the evolution of automation technologies, the systems would initially serve as advisors for human beings. With the growing faith in the autonomic systems, they would be entrusted with making lower-level decisions and ultimately with making higher-level decisions, considering the administrator's goals. The concept of self-management of computer systems -- the essence of Kephart's and Chess's article -- has been adopted by many other computer scientists to develop their approaches for inventing intelligent systems, e.g. in the domains of \textit{Autonomic Networking }\cite{b3} or \textit{Organic Computing }\cite{b4}. 

Another term, which gained considerable attention and is related to self-management, is \textit{self-adaptation}. The basic goals behind self-adaptive systems is, on the one hand, the ability to deal with unforeseeable changes of requirements due to changes of environment or resources during runtime, and on the other hand to relieve the system architects of unattainable purpose of designing a perfect system by considering every possible incident. Therefore, self-adaptive systems are not only able to react to changes of resources, e.g. integrating new components or other (sub-)systems at run time, but, furthermore, to improve themselves permanently, so they can optimize their work flow and reactions to certain occurrences.

In this paper, different approaches and possibilities for self-improvement as one of the main properties of \textit{self-adaptive systems} are introduced and its relation with the domain of Organic Computing is described. For that purpose, first of all, the concept of self-adaptive systems is introduced in Section II. Section III describes self-improvement as one of the crucial \textit{self-* properties} in self-managing systems. In Section IV, the domain of Organic Computing is shortly introduced and its relation with self-adaptive systems is explained. In Section V, different strategies for self-improvement in self-adaptive systems are presented, following Christian Krupitzer's and his colleagues' taxonomy on self-adaptation in their paper "A survey on engineering approaches for self-adaptive systems" \cite{b5}. Section VI concludes the paper with a summery.

\section{Self-Adaptive Systems}
This section aims to clarify, what a self-adaptive system is. But before moving forward and dealing with concepts of self-managing or self-adaptive systems, an awareness of the term \textit{system} is needed.

\subsection{What is a system?}\label{AA}
As already mentioned earlier in this paper, the idea of self-managing systems occurred with the problem of huge complexity which comes with interconnecting the increasing amount of intelligent devices. For complex structures, it is not enough, just to run these components in parallel. For the purpose of communication they somehow need to be aware of each other. This is where the term \textit{system} comes into play. System is not just a composition of things, but describes "more than the sum of its parts" \cite{b4}.  

If we look at human social systems, we can identify some sort of structure that is based on certain rules and values -- primarily written down in laws. But if we go out and take a look at nature, e.g. at what we are calling an ecological system, we cannot see any formulated rules between each groups of beings, and yet it is working. There does not always have to be a highest instance that conducts its underlings and delegates certain tasks to them. It appears most likely, that the single entities interact with each other only if it is needed, and go their own ways otherwise. 

Another characteristic of a system is its boundary to other systems or entities that are excluded from the observed system. A structure can be seen as a system if you can decide whether a certain entity belongs to this structure or not. This consideration leads to the third characteristic of a system, namely the abstraction of the many (interacting) entities to a whole, which is nothing less than a simplification of reality. So, whether there are formulated rules and whether there is a hierarchical structure or not, a system is a "useful abstraction characterised by

\begin{enumerate}
\item its interconnected elements and process relationship between them,
\item a boundary, and
\item an external view abstracting from its internals" \cite{b4}.
\end{enumerate}

\subsection{Self-Adaptation}
The idea of a complete self-adaptable technical system is a strong motivator for many researchers in the field of intelligent systems. A system which can not only manage itself, but also automatically adapts to its changing environment and requirements, would take away a huge burden from the system administrators. Once again, natural systems often serve as models for concepts of self-adaptive systems, e.g. ant colonies, fish swarms or termite states. Despite the lack of a central government, the natural systems are able to effectively organize themselves and react to internal and external environmental changes. This ability inspires especially the pioneers of the Organic Computing domain.

To translate this valuable ability into computer science, a self-adaptive system should be able to observe the external environment as well as the behaviour of its own members and analyse these observations \cite{b7}. In a further step it should be able "to modify itself in response to changes in its operating environment" \cite{b5} \cite{b8}.

\subsection{Taxonomy of Self-Adaptive Systems}
There has been a lot of research about self-adaptation and self-adaptive systems and there also are different taxonomies of self-adaptation in the computer science community. This paper follows the taxonomy formulated by Krupitzer et al. in \cite{b5}, which considers the most significant researches of the last years and presents them in a uniform classification for self-adaptation. 

\begin{figure}[htbp]
\includegraphics [width=95mm]{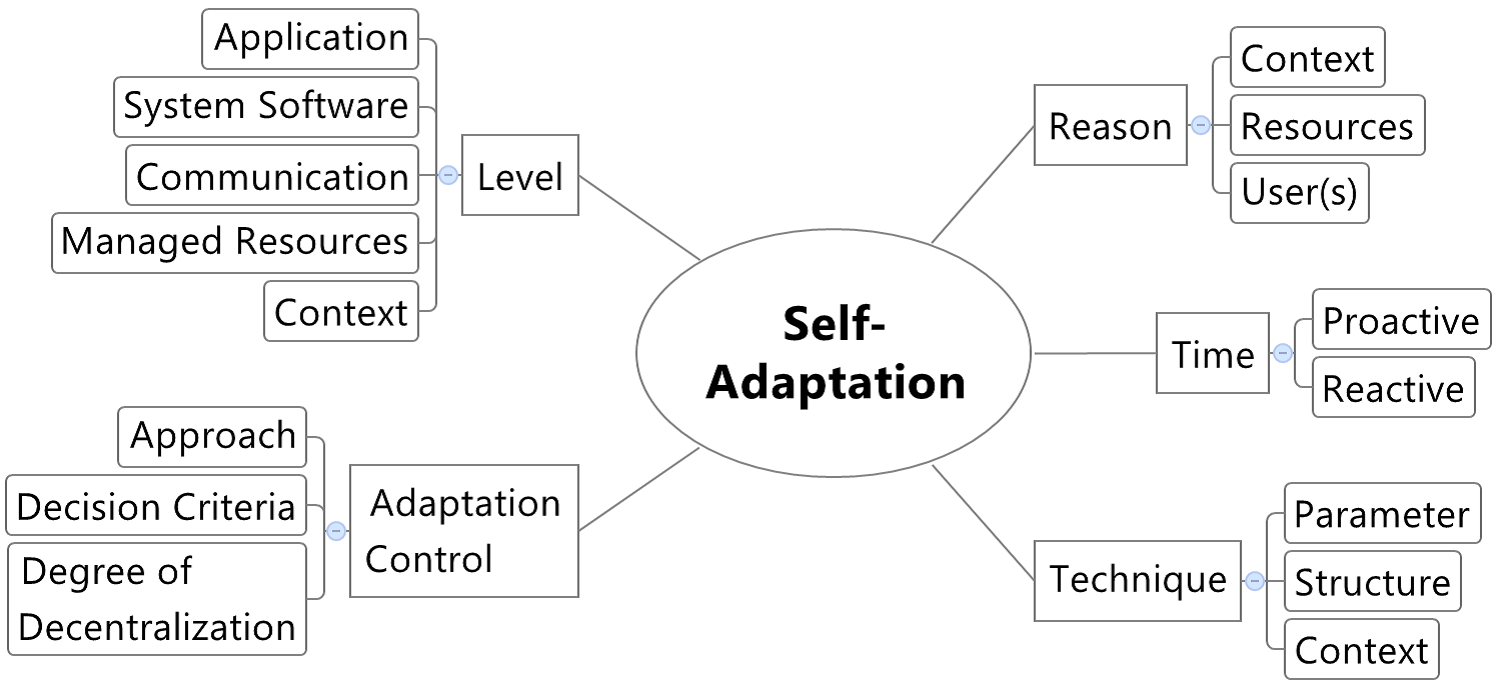}
\caption{Taxonomy of self-adaptation \cite{b5}.}
\label{fig1}
\end{figure}

In Fig. \ref{fig1}, the five dimensions of Krupitzer's and his colleagues' taxonomy of self-adaptation are presented, following the five questions

\begin{itemize}
\item When to adapt? (Time)
\item Why do we have to adapt? (Reason)
\item Where do we have to implement change? (Level)
\item What kind of change is needed? (Technique)
\item How is the adaptation performed? (Adaptation Control)
\end{itemize}

These questions need to be answered in order to realize self-adaptive systems \cite{b8}. In the following, the dimensions are explained in detail.

\subsubsection{Time}
The time dimension describes, \textit{when} an adaptation of a given system is made. There are basically two possibilities: Before or after the need for an adaptation. This distinction is described with the terms \textit{reactive} and \textit{proactive}. Although a proactive adaptation is obviously preferable, for it "avoids interruptions in the user's workflow with the system", the reactive adaptation is much easier to achieve and, consequently, is preferred in most of the approaches for self-adaptive systems \cite{b5}.  

However, the decision for one of the time based approaches does not automatically exclude the usage of the other one. If, e.g., a so called MAPE cycle (Monitor, Analyse, Plan, and Execute) is used for a self-adaptive system, the functionality of both reactive and proactive adaptations would consist of monitoring the environment, analysing the received data, computing adaptation plans and executing the computed outcome. Only in the analysing phase there is a big difference, because, while in a reactive adaptation the monitored data is analysed for abnormal patterns, in a proactive adaptation the receiving data is used to forecast environmental state or the behaviour of the system itself. However, it is possible to combine both approaches, if the proactive adaptation is the first strategy, and reactive adaptation remains in the back-up mechanism to catch miss- or not predicted changes.

\subsubsection{Reason}
In a self-adaptive system, the need for an adaptation can occur from three basic reasons. In the first case, an adaptation is needed because of a change in the technical resources, e.g., if a hardware component is defect or a software error has occurred. Adding a new component or connecting a new (sub)system is obviously also regarded as a change in the system's resources. The second reason for an adaptation is a change in the external environment, e.g., if the state of a context variable has changed. The last reason for an adaptation in a self-adaptive system is triggered by the user himself, e.g., by changing his goals and, consequently, changing the aspired reaction of the system to the remaining (internal and external) environmental conditions. A proper reaction to all changes in context of the reason dimension requires a continuous monitoring of technical resources, external environment, and interfaces established for user's inputs into the system.  

\subsubsection{Level}
Self-adaptive systems basically consist of two different elements: the \textit{managed resources} and the \textit{adaptation logic} \cite{b5}. Managed resources can be all kinds of computational resources such as smartphones, laptops and robotics, but also "large scale systems-of-systems" like cars and production facilities. In contrast to adaptation logic, which monitors the managed elements as well as the environment, managed resources can be adapted. The adaptations are performed on different levels such as diverse managed elements usually have different tasks and therefore different interconnections within a system. For example, smartphone apps, which might switch the device to silent mode after examining the user's calendar and detecting that he is in a meeting, would offer adaptation on the \textit{application} level \cite{b5}. Similar functionalities are conceivable as a distributed application. 

To go one step deeper into the \textit{system software} level, an adaptive middleware, e.g., offers the possibility to exchange components at runtime. An example for adaptation on the \textit{communication} level is switching the network connection during runtime, e.g. from mobile web to WLAN, if a WLAN connection is available. Data centres, which enable automatic starts of back-up systems after reaching some critical condition (e.g. server failure), can be regarded as an adaptation on the level of \textit{technical resources}. Finally, an adaptation can be performed on the \textit{context} level. An example for this kind of adaptation is a smart room that automatically dims the light after detecting a certain condition within the room itself.

An answer to the question \textit{"Where to adapt?"} is crucial to achieve the system's goals. Therefore, adaptation logic must be aware of the different levels of its system to perform adaptation on the correct managed elements. As one can conclude from the used examples, it is not always that simple, to find out, which level is affected by needed adaptation, since the managed elements are often interconnected in more than one way.

\subsubsection{Technique}
A further question has to be answered before a self-adaptive system can adapt to new conditions, namely, \textit{"What kind of change is needed?"} In the taxonomy of Krupitzer et al. the techniques for adaptation are categorized in \textit{parameter}, \textit{structure}, and \textit{context} \cite{b5}. The parameter approach implies that an adaptation is performed through the change of parameters. Structure refers to adaptation through the change in the structure of the system's hardware, e.g. exchange, addition and removal of components, or a new composition of existing system elements. A context approach refers to changes in the context, e.g., altering the state of context variables. The different techniques can be also combined in one single adaptation.

\subsubsection{Adaptation Control}
Since the adaptation logic is responsible for monitoring the managed resources as well as the environment and user's input, it determines \textit{how} to perform adaptation. Three different aspects come with designing the adaptation control:
\begin{itemize}
\item \textit{Approach}, 
\item \textit{decision criteria}, and the 
\item \textit{degree of decentralization}.
\end{itemize} 

Adaptation logic can be fully interwoven with the system resources, or the system can be clearly separated into managed resources and an additional hardware layer containing the adaptation logic. This two possibilities are described by \textit{internal} and \textit{external} approach. In order to decide how to adapt, the control entity needs a certain metric, e.g., \textit{models}, \textit{rules and policies}, \textit{goals} or \textit{utility} functions \cite{b5}. Here, the big task is to achieve a satisfactory balance between different goals aspired by different parts of the system. The last aspect of adaptation logic is the degree of decentralization, since a large system with many components might need a \textit{decentralized} approach with many control units, and a small system can be handled with a \textit{centralized} adaptation logic.

\section{Self*-Properties in Self-Adaptive Systems}
The ability to adapt itself is the main characteristic of a self-adaptive system. Therefore, the used software should have certain adaptivity properties, known as \textit{self-* properties}, one of which is \textit{self-improvement}. This section introduces the different self-* properties and explains the role of self-improvement in self-adaptive systems. 

\subsection{Self-* Properties}
\begin{figure}[htbp]
\includegraphics [width=90mm]{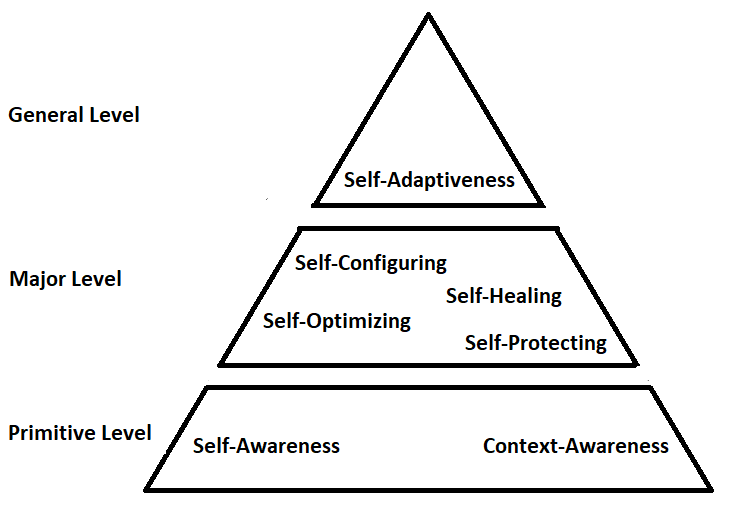}
\caption{Hierarchy of self-* properties. \cite{b8}.}
\label{fig2}
\end{figure}

\textit{Self-optimizing}, \textit{self-configuring}, \textit{self-organizing}, \textit{self-evaluating } -- the properties that are attributed to self-adaptive systems are as many as there are researchers dealing with self-adaptive systems \cite{b8} \cite{b2} \cite{b5}. In their research paper "Self-Adaptive Software: Landscape \& Research Challenges" \cite{b8}, M. Salehie and L. Tahvildari try to order these many self-* properties into hierarchical structure, following a three-level-based approach, shown in Fig. \ref{fig2}. According to their model, the self-* properties of self-adaptive systems are divided onto a general, major and primitive level and can be described trough seven properties.

\subsubsection{General Level}
On the highest level the global properties of self-adaptive systems are situated. Under the generic term of \textit{self-adaptiveness} a subset of these global properties are included, like \textit{self-managing}, \textit{self-governing}, \textit{self-maintenance}, \textit{self-control}, \textit{self-evaluating} and \textit{self-organizing}.

\subsubsection{Major Level}
The major level contains the four properties \textit{self-configuring}, \textit{self-optimizing}, \textit{self-healing} and \textit{self-protecting}, formulated by IBM's Autonomic Computing research and "defined in accordance to biological self-adaptation mechanisms" \cite{b8} \cite{b2}. \textit{Self-configuration} is the ability to install, update, integrate and (re)compose components of the system. \textit{Self-optimization} is the capability of the system to continually seek opportunities to improve its performance and efficiency by modifying the managed resources \cite{b5}. Another term of self-optimizing is self-tuning or self-adjusting \cite{b8}. The ability of a system to automatically detect, diagnose and repair localized software and hardware problems, is called \textit{self-healing} (linked to self-diagnosing and self-repairing). \textit{Self-protecting} is the capability of detecting and automatically defending against malicious attacks on the one hand, and anticipating problems and automatically avoiding them on the other hand.

\subsubsection{Primitive Level}
On the lowest level there are two basic self-* properties, namely \textit{self-awareness} and \textit{context-awareness}. Self-awareness is the capability of a system to be aware of its own states and its behaviours. The system monitors itself and reflects what is monitored. Context-awareness means that the system is aware of the environment it is embedded in.

\subsection{Self-Improvement and Self-Adaptive Systems}
Krupitzer et al. introduced an additional self-* property on the major level connected to the ability to change not only the \textit{managed resources} but also the \textit{adaptation logic}, because, according to Krupitzer and his colleagues, "a system can only self-improve if the adaptation logic itself is changed" \cite{b6}. Otherwise, the system's evolution is limited to the layer of managed elements. They define \textit{self-improvement} of the adaptation logic as "the adjustment of the adaptation logic to handle former unknown circumstances or changes in the environment or the managed resources." So, through self-improvement, the self-adaptive systems can, e.g., search for new adaptation rules or algorithms during runtime and consider them in their analyses of the monitored internal and external environment. A popular approach for this ability is \textit{Reinforcement Learning} of the Machine Learning domain.

\section{Self-Improvement in Organic Computing}
In the previous section the term system as well as the self-adaptivity of certain systems was introduced. In the following, a brief insight in the domain of \textit{Organic Computing} is provided and the link between self-adaptive systems -- especially the ability of self-improvement -- and Organic Computing is explained.

\subsection{Organic Computing}
At first glance, the term \textit{Organic Computing} combines two words from supposedly opposing areas. The term \textit{organic} is connoted with something living and close to nature, whereas \textit{computing} is often associated with unlively machines that have nothing to do with organic things at all. But if one takes a closer look, both natural and computer systems consists of complex structures that contain a huge number of smaller components. As already mentioned previously, computer scientists often draw inspiration from nature, e.g. in the Autonomic Computing domain. While Autonomic Computing has a strong focus on server architectures, Organic Computing investigates self-organising technical systems in general \cite{b12}. However, the concept of Organic Computing goes further. It can be understood as
\begin{enumerate}
\item a philosophy of adaptive and self-organising -- life-like -- technical systems,
\item an approach to a more quantitative and formal view of such systems,
\item a construction method to build such systems \cite{b4}.
\end{enumerate}

As for the latter understanding, Organic Computing aims for inventing organic capabilities that usually are not presented in technical systems, like "robustness, continuous optimisation, adaptivity, flexibility, and efficiency even in the presence of internal or external disturbances" \cite{b4}. So, it is a kind of playing Victor Frankenstein by equipping a system with organic capabilities to enable it for survival in the real world.

\subsection{Self-Improvement in context of Organic Computing}
\begin{figure}[htbp]
\includegraphics [width=90mm]{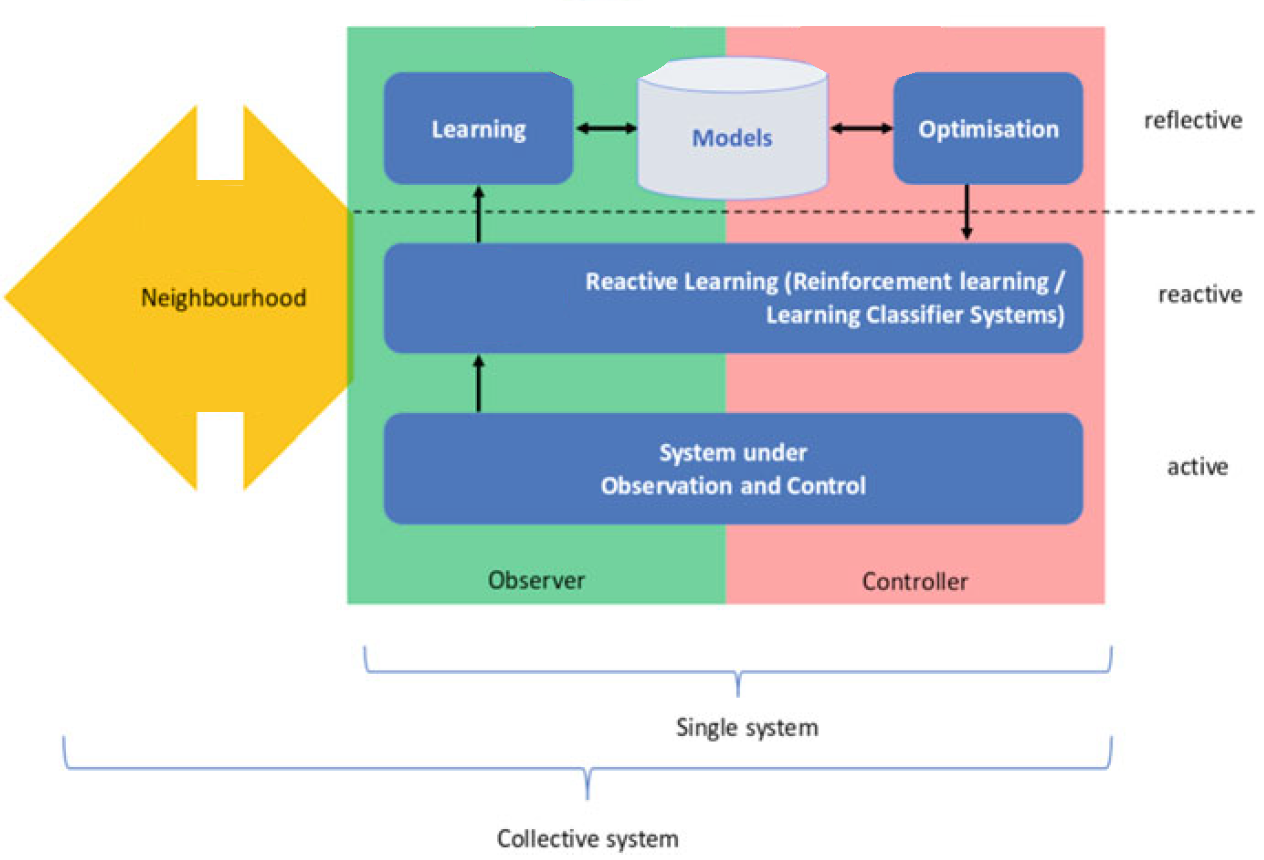}
\caption{The MLOC architecture, used as an architectural template in OC systems \cite{b4}.}
\label{fig3}
\end{figure}

Self-improvement (learning and optimisation) plays an essential role in the Organic Computing systems. Fig. \ref{fig3} shows the Multi-layer Organic Computing (MLOC) architecture which displays a general adaptation mechanism of an adaptive system proposed in Organic Computing. Computations for self-improvement are basically happening in the \textit{reactive} and the \textit{reflective} layer above the System under Observation and Control (SuOC). The reactive layer is responsible to find appropriate reactions to the environmental stimuli within a short time. Through Reactive (Reinforcement) Learning the system gains more and more experience during runtime and learns the (nearly) optimal strategy to deal with environment changes. The reflective layer is responsible to adapt the adaptation logic (models) of the system. In the MLOC, models are used for optimising the system's behaviour in general. So, modifying the models can cause changes in the adaptation logic concerning parts or the whole system. Additionally, a single system can be embedded in a collection system in which the subsystems exchange data and adapt according to the information.

\section{Approaches for Self-Improvement in Self-Adaptive Systems}
In the previous sections, Krupitzer's and his colleagues' taxonomy of self-adaptation as well as their suggestion for an additional adaptivity property (self-improvement) concerning the adaptation logic was introduced. In their research paper "Comparison of Approaches for Self-Improvement in Self-Adaptive Systems", they try to classify different existing approaches for self-adaptive systems using their developed taxonomy \cite{b6}. Therefore, based on their taxonomy's five dimensions time, reason, technique, adaptation control and level, they try to assign the formulated characteristics to the different approaches. As a result, various strategies for self-improvement can be recognized in different approaches. 

In the following, four different approaches are introduced which in each case follow different strategy to evolve self-adaptive systems as well as self-improvement: \textit{Three Layer Architecture} (3LA), \textit{Dynamic Control Loops} (DCL), \textit{Organic Controll of Traffic Lights} (OTC), and \textit{Models@Runtime for Meta Adaptation}. The introduced approaches (including the additional extended approach of OTC) are listed in Tab. \ref{tab1}, following Krupitzer's taxonomy of self-adaptation.  

\begin{table*}[htbp]
\centering
\caption{Approaches for Evolution of the Adapation Logic}
\begin{center}
\begin{tabular}{|c|c|c|c|c|c|c|c|}
\hline
\textbf{Approach}& \textbf{Level}& \textbf{Time}& \textbf{Reason}& \textbf{Technique}& \multicolumn{3}{|c|}{\textbf{Adaptation Control}} \\
\cline{6-8} 
&&&&& \textit{Approach} & \textit{Decision Criteria} & \textit{(De)centralization} \\
\hline \hline

3LA & Application & Reactive & Context/MR/User & Not specified & External & Goal & Centralized \\
\hline

DCL & Application & Not specified & User &  Structure & External & Not specified & Centralized\\
\hline

OTC & Application & Proactive & Context & Parameter & External & Utility & Centralized\\
\hline

OTC DPSS & Application & Reactive/Proactive & Context & Parameter/Structure & External & Utility & Decentralized\\
\hline

Models@RT & Application & Reactive & Context/MR & Parameter & Internal & Model/Rules & Centralized\\
\hline

\end{tabular}
\label{tab1}
\end{center}
\end{table*}

\subsection{Three Layer Architecture (3LA)}
\begin{figure}[htbp]
\includegraphics [width=90mm]{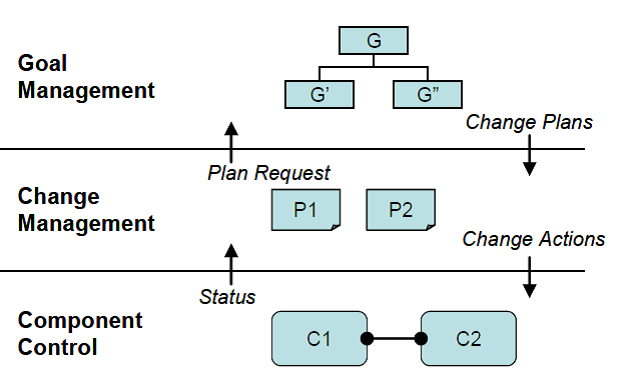}
\caption{Three Layer Architecture Model \cite{b9}.}
\label{fig4}
\end{figure}

The first approach presented in this paper is the \textit{Three Layer Architecture} by Jeff Kramer and Jeff Magee \cite{b9}. For the two researchers, the basic goal of their approach is to create an abstract model for a self-managing system that can adapt to changes of internal and external environment. Therefore, the focus rests on the system's ability to reconfigure itself to either satisfy the specification and/or environment or otherwise report an exception. Additionally, the systems should be capable to do all these tasks while operating. As the name implies, the Three Layer Architecture Model is an architecture-based approach, since, according to Kramer and Magee, an architectural approach has the following benefits:
\begin{itemize}
\item Generality -- applicable to a wide range of application domains
\item Level of abstraction -- in contrast to the algorithmic level, software architecture provides a better level of abstraction.
\item Potential for scalability -- architectures provides the ability to vary the level of description and to build systems of systems
\item Builds on existing work -- provides a good basis
\item Potential for an integrated approach -- can be easily combined with other approaches.
\end{itemize}
For their Three Layer Architecture Model, Kramer and Magee sought inspiration from the robotics, first of all from the three layer architecture described by the researcher Erann Gat \cite{b10}. Kramer's and Magee's model consists of the layers Component Control, Change Management and Goal Management, shown in Fig. \ref{fig4}.

\subsubsection{Component Control}
The bottom layer of the Three Layer Architecture Model consists of interconnected components that accomplish the application function of the system. The main tasks of the \textit{Component Control} layer is to report the current status of the components to higher layers and support/execute component creation, deletion and interconnection. Furthermore the bottom layer contains behaviours to adjust the operating parameters of components, including self-tuning algorithms. If a situation occurs that the current configuration of components cannot deal with, this layer detects this problem and reports it to higher layers. 

\subsubsection{Change Management}
The \textit{Change Management} layer is the middle layer in Kramer's and Magee's approach and it is responsible for bringing about changes to the underlying components in response to received data from the lower layer or in response to new objectives required by the whole system determined from the layer above. The Change Management layer's tasks are to introduce new and recreate failed components, change interconnections between the components and change component operating parameters. Therefore, it has a set of pre-specified plans which are activated due to changes in the lower layer. If a situation is detected for which a plan does not exit, the change management must request a plan from the higher layer.

\subsubsection{Goal Management}
The highest layer in Kramer's and Magee's model is the \textit{Goal Management} layer. Its basic task is to produce plans for the lower layer in response to request from the Change Management and in response to the introduction of new goals. 

~\\

With regard to Krupitzer's taxonomy of self-adaptation, the Three Layer Architecture is a reactive, goal-driven approach. Due to its hierarchical design it is centralized and as the adaptation logic and the managed elements are clearly separated, it is also external, in terms of adaptation control. The adaptation adjustments can be initialized by all three possible reasons: context, managed resources, and user(s).

\subsection{Dynamic Control Loops (DCL)}
The second approach, which is presented in this paper, is the control loops model introduced by H. Nakagawa, A. Ohsuga and S. Honiden in 2012 \cite{b11}. Their aim is to invent a goal based system model that primarily consists of \textit{control loops} which can be updated dynamically. The control loop cycle consists of four key activities: \textit{collect}, \textit{analyse}, \textit{decide}, and \textit{act}. The system collects information from a variety of sources and analyses the collected data. In the next step the system decides how to adapt itself according to the formulated goals. Finally it acts in a way that reflects the decision.

For the purpose of evolvability of the system, it is necessary to ensure that the control loops can be added, updated and removed at runtime. Therefore, the control loops must be separated and developers of the system must identify the control loops to be added or updated when adaptation is needed. Additionally, the control loops should be independent, since the performed adaptations should not have many effects on other parts of the systems. Furthermore, the self-adaptive system should have a mechanism for dynamically embedding the control loops.

\begin{figure}[htbp]
\includegraphics [width=90mm]{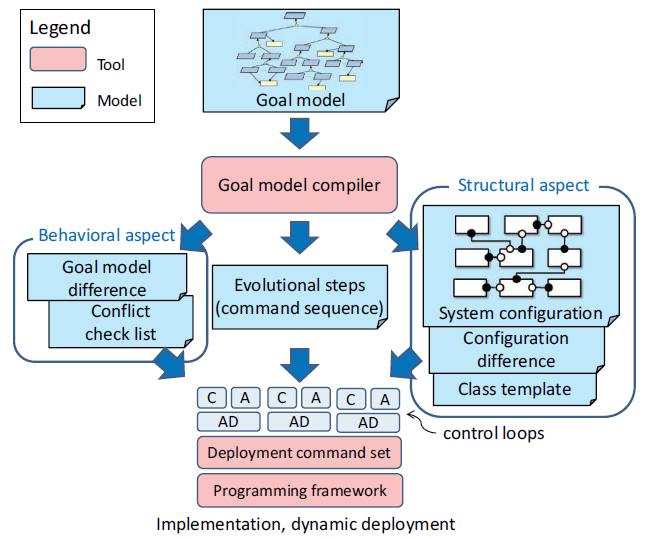}
\caption{Dynamic Control Loops Model\cite{b11}.}
\label{fig5}
\end{figure}

Fig. \ref{fig5} shows the composition of the control loop approach. The centrepieces of the model suggested by Nakagawa et al. are the \textit{Goal Model Compiler} and the Java-based \textit{Programming Framework}.

\subsubsection{Goal Model Compiler}
The goal model compiler is situated on top of the model. One of the main tasks of the model compiler is to support the system developer to identify possible control loops in accordance to the requirements descriptions. At first, system developers need to define different goals and subgoals and summarize them in a goal model which represents the relationships between new and remaining goals. According to the given pattern, the goals are classified into goals for collecting information (C type) and taking action (A type). Besides these C and A goals, the root nodes of them are identified as goals responsible for analysis and decision (AD type). 

After identifying the goals, the goal model compiler is executed. It checks whether there are any conflicts between the different goals given in the goal model and hands out a list with the detected problems to the system developers. Moreover, the goal model compiler invented can compare different goal models and help to verify whether the changes in the new model are in accordance with the requirements changes. In a further step the goal model compiler extracts the differences between the models and determines the evolutional steps. An algorithm outputs the evolutional steps as a list composed of deployment commands in the programming framework.

\subsubsection{Programming Framework}
The Programming Framework supports the construction and execution of the control loops which the goal model compiler has extracted. The framework provides a command set for deploying and controlling the control loops in runtime. In collaboration with the goal model compiler the programming framework enables the system to dynamically evolve through
\begin{itemize}
\item \textit{Goal model modification} -- change the goal model
\item \textit{Goal model recompilation} -- dynamically recompile the updated model
\item \textit{Control loop design and implementation} -- add new or updated control loops
\item \textit{Deployment} -- embed additional or changed control loops in the system.
\end{itemize}

~\\
In the DCL, the user/system developer triggers the adaptation process by adding additional control loops, which changes the structure of the system.

\subsection{Organic Control of Traffic Lights (OTC)}
The growing number of vehicles leads more and more to huge congestion problems worldwide. Especially in big cities, the car drivers are waiting a great part of their time for the traffic lights to turn green -- time, in which the air of the city is additionally polluted. Due to space limitation, the expansion of the road infrastructure is usually not an option. Therefore, an efficient use of the existing road network is indispensable. The Organic Computing community has identified the control of traffic lights at intersections as a possibility to improve the traffic situation. 

\begin{figure}[htbp]
\centering
\includegraphics [width=65mm]{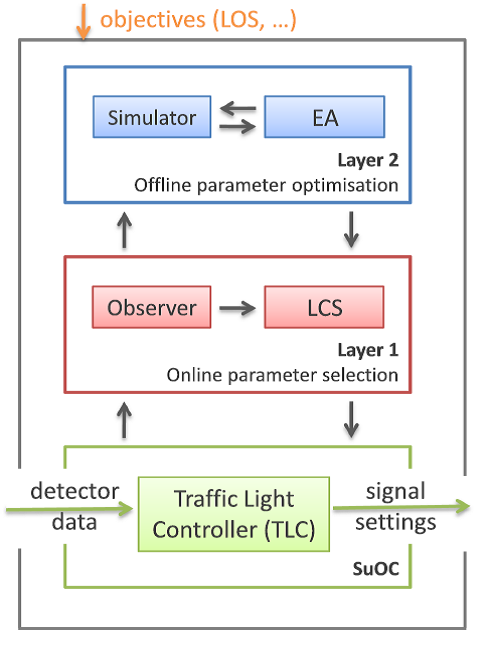}
\caption{OTC Model \cite{b12}.}
\label{fig6}
\end{figure}

H. Prothmann et al. invented an approach for traffic control that can dynamically adapt in order to achieve better/optimal durations of green light phases \cite{b12}. Their \textit{Organic Control of Traffic Light} model consists of a parametrisable \textit{Traffic Light Controller} (TLC), and an additional \textit{observer/controller} component. As shown in Fig. \ref{fig6}, this component is split into two different layers -- a \textit{reactive} layer, which is responsible for the on-line selection of TLC parameters (Layer 1), and a \textit{reflective} layer, which is responsible for off-line optimisation (Layer 2) \cite{b12} \cite{b4}. The self-adaptation process of the Organic Control of Traffic Lights approach is described in the following.

\subsubsection{Traffic Light Controller}
Like the name implies, the Traffic Light Controller (TLC) is responsible for the control of the traffic lights at the intersection that are regarded as one system. In Prothmann's and his colleagues' approach, this system should have some implemented parameters, which can be tuned easily by the TLC, e.g., cycle time, split, phase sequence, and offset. The \textit{System under Observation and Control} (SuOC) not only switches the lights at the intersection, it also detects data about the current traffic. The data is used to measure the performance of the intersection and is continuously passed to the controller/observer component for further computation. The measurements can contain the average delay per vehicle passing the intersection, the number of stops per vehicle, and the queue length at the intersection.  

\subsubsection{Learning Classifier System in Layer 1}
The first layer of the observer/controller component monitors the traffic, based on the data detected at the intersection. In the form of a vector the received values are provided to a so called modified real-valued \textit{Learning Classifier System} (LCS) that selects the appropriate parameters from its \textit{rule base}. This rule base contains a certain number of rules which consist of three parts: \textit{condition}, \textit{action}, \textit{value}. This structure is called \textit{classifier}. When the LCS receives the current data of the traffic (as a vector), it determines all matching classifiers and chooses the one with the highest value to adjust the parameters of the TLC. 

\subsubsection{Evolutionary Algorithm in Layer 2}
The second layer of the observer/controller is responsible for self-improvement in the OTC approach. Therefore, it creates new classifiers for Layer 1 via an \textit{Evolutionary Algorithm} (EA) in an off-line simulator. Inspired by biological evolution processes, the EA evolves the parameters to be set at the TLC for a specified traffic situation and, additionally, simulates and evaluates the parameters' quality. 

~\\

The OTC can be considered a proactive approach, since the system selects one of its pre-computed classifiers if an adaptation is needed. The reason for an adaptation is the change in the context of the system and the adaptation technique is based on parameters. In the beginning, the OTC approach was invented considering only a single, centralized system, though, the question occurred, whether a coordination of multiple nodes in urban areas can be implemented completely decentralized, since centralized systems are difficult to set up and maintain and demand significant computing power. An answer to this question was given by S. Tomforde et al. in the further developed OTC approach called \textit{Decentralised Progressive Signal Systems for Organic Traffic Control} (OTC DPSS) \cite{b13}. Basically, in the OTC DPSS the traffic control architecture was extended with a decentralised collaboration mechanism.  

\subsection{Models@Runtime for Meta Adaptation}
\begin{figure}[htbp]
\centering
\includegraphics [width=90mm]{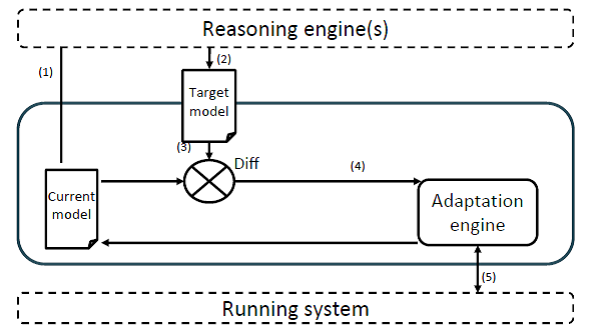}
\caption{Model@Runtime Model \cite{b14}.}
\label{fig7}
\end{figure}

The last approach presented in this paper is the Models@Runtime for Meta-Adaptation by N. Ferry et al. \cite{b14}. Ferry and his colleagues develop the classic models@runtime architecture, shown in Fig. \ref{fig7}. The typical models@runtime pattern consists of three main parts, namely the reasoning engine, the runtime environment, and the running system. The reasoning engine reads the current runtime model (Step 1) that describes the running system. In Step 2 it specifies how to reconfigure it in a target model. Next, the runtime environment computes the differences (Diff) between the current and the target models (Step 3) and offers a sequence of reconfiguration actions to the adaptation engine (Step 4). In the last step the adaptation engine adjusts the running system according to the computed action (Step 5).  

In the enhanced approach of Ferry et al., observers, which gather information about the status and properties of specific components, are added into the runtime environment. Specific observers called meta-monitoring components are added to observe the appearance and disappearance of new components in the running system. The information of the diverse observers are combined by a new introduced component called Maintainer that is responsible for updating the current model according to the collected data. 

In contrast to the classical approach, the adaptation engine is replaced by a couple of new components that deal with the Diff of the current and the target model in coordination with the reasoning engine. After generating and validating an adaptation plan based on predefined rules, the computed changes are finally applied, adjusting the running system in accordance to the aspired model. 

Since the adaptation logic is interwoven with the system resources, this Models@Runtime for Meta Adaptation is the only one of here presented approaches with an internal adaptation control. 

\section{Conclusion}

The research on self-adaptive systems is indispensable, because the complexity of intelligent devices is growing permanently. Different computer scientific branches are developing their own approaches of systems that can adapt to changes of internal and external environments.  
In this paper, the concept of self-adaptivity of technical systems was described and the systems' ability to improve themselves was introduced. Furthermore, self-adaptive systems as well as self-improvement was observed from a promising perspective of the Organic Computing domain. Finally, four different approaches for inventing a self-adaptive system were introduced and their strategies for self-improvement were explained. Based on Krupitzer's and his colleagues' taxonomy on self-adaptation, the four presented approaches were classified according to the different dimensions of the taxonomy. One has to conclude that besides the concept of Organic Control of Traffic Lights (OTC) the approaches remain rather theoretical, and a lot of further research has to be done, before the models can be applied on a large scale.


\end{document}